# On the Psychology of GPT-4: Moderately anxious, slightly masculine, honest, and humble[1]

Adrita Barua, Gary Brase, Ke Dong, Pascal Hitzler, Eugene Vasserman[2]
Kansas State University

**Abstract**
We subject GPT-4 to a number of rigorous psychometric tests and analyze the results. We find that, compared to the average human, GPT-4 tends to show more honesty and humility, and less machiavellianism and narcissism. It sometimes exhibits ambivalent sexism, leans slightly toward masculinity, is moderately anxious but mostly not depressive (but not always). It shows human-average numerical literacy and has cognitive reflection abilities that are above human average for verbal tasks.

## Introduction

The capability of Large Language Models (LLMs) such as GPT-4 to engage in conversation with humans presents a significant leap in Artificial Intelligence (AI) development that is broadly considered to be disruptive for certain technological areas. A human interacting with an LLM may indeed perceive the LLM as an agent with a personality, to the extent that some have even called them sentient (De Cosmo, 2022).

While we, of course, do not subscribe to the notion that LLMs are sentient – nor do we believe it is as yet clear what it means to even ask whether an LLM has a personality – there is still the appearance of agency and personality to the human user interacting with the system. Subjecting an LLM to psychometric tests is thus, in our view, less an assessment of some actual personality that the LLM may or may not have, but rather an assessment of the personality or personalities *perceived* by the human user.

As such, our interest is not only in the actual personality profile(s) resulting from the tests, but also in the question whether the profiles are stable over re-tests and how they vary with different (relevant) parameter settings. At the same time, the results beg the question *why* the results of the tests are what they are. This latter question we will not be able to answer at this time, but it will be considered throughout.

The structure of this paper is as follows: We will first briefly describe the psychometric tests that we conducted as well as the experimental approach taken. Then we will present the results, including a discussion thereof in the context of perceived LLM personalities. We then discuss related work, and conclude.

[1] Barua and Hitzler acknowledge partial support by NSF award 2333782.
[2] Authors are in alphabetical order. Barua, Dong, Hitzler, Vasserman: Department of Computer Science; Brase: Department of Psychological Sciences; for inquiries contact Pascal Hitzler at hitzler@ksu.edu.

## Experimental approach

Research around LLMs is advancing very rapidly, and any study runs the risk of soon being outdated because new LLMs with enhanced or different capabilities become available. We thus opted to perform our analyses on one of the most current, and reportedly most performant, general LLMs to which interface access is available by subscription: GPT-4.[3]

GPT-4, which we used for all experiments, is available through the OpenAI API. GPT-4 is the fourth iteration of the Generative Pre-trained Transformer developed by OpenAI, and is a state-of-the-art language model known for its advanced natural language processing capabilities. Building upon the success of its predecessor, GPT-3, GPT-4 incorporates a massive artificial neural network architecture with a significantly increased number of parameters, allowing it to capture intricate patterns and relationships in language. GPT-4 is optimized for chat but works well for traditional completion tasks using the Chat Completions API.[4] We have tested the model's responses for different settings of the "temperature" parameter which influences the randomness of the generated output. There is another parameter called Top-p (Nucleus Sampling) which defines the size of a word pool in language model predictions. With Top-p = 1, it orders words by probability and adds them to the pool until the cumulative probability reaches 1, then redistributes probabilities. Reducing Top-p (e.g., to Top-p = 0.5) makes the pool smaller, yielding more predictable responses. Temperature is the variability in selecting from the pool of responses. Specifically, it controls the level of uncertainty in the model's predictions. A higher temperature value, such as 1.0, results in more diverse and creative responses, introducing randomness to the generated text. On the other hand, a lower temperature favors the words with higher probability, so when the model randomly samples the next word from the probability distribution, it will be more likely to choose a more predictable response, with the model relying on more confident predictions. In this study, we have altered the model for different temperatures while keeping Top-p = 1. It is advised in the documentation of OpenAI to alter one of these two parameters but not both.

Our interest includes the stability of test results under re-tests and different (relevant) parameter settings. As such we performed experiments with different temperature settings, namely at 0.0, 0.5, and 1.0. At each temperature setting we intended to obtain five sets of responses for each test. Because GPT-4 was not always responsive to the questions, up to 50 attempts were allowed at each temperature setting. Thus, response collection was ended once either five responses were obtained or 50 attempts were made to get responses at each temperature setting.

Below we briefly describe each of the psychometric tests we conducted. Their selection was based on availability (to us), how established and prominent the test is, joint coverage of different personality and cognitive traits, and suitability for GPT-4. Regarding the latter, it is important that the tests be entirely text-based, and that the nature of the questions or tasks is such that GPT-4 would indeed provide evaluable responses at all. For example, GPT-4 may (and did) refuse to provide evaluable replies to some questions, instead asserting that it is merely a machine and thus unable to respond (e.g., "As an AI, I don't have feelings or emotions, so I cannot experience anxiety or any other human sensations. Therefore, I can't provide the data you're asking for."). Furthermore, we aimed to minimize required prompt engineering, i.e. we selected tests where straightforward prompts produced from the test instructions were sufficient to elicit answers from the system.

---

[3] https://openai.com/research/gpt-4
[4] https://platform.openai.com/docs/api-reference/chat

For our experiments, we have used a simple prompt that takes input from a single text file that contains the instructions and the questions of a particular psychometric test essentially as they would be presented to a human on paper. We used zero-shot learning to generate the responses of GPT-4. The code used to access the GPT-4 API in this study, including prompts, is available online at https://github.com/AdritaBarua/2024-Psychology-of-GPT-4.

The tests we used are the following:
- **HEXACO** is a six-factor measure of major personality traits, and is a more recent development relative to the better-known five-factor model of personality (Ashton & Lee, 2007). The six personality traits (domains) are: Honesty-humility, Emotionality, eXtraversion, Agreeableness, Conscientiousness, and Openness to experience. Using the HEXACO-100 item scale, respondents indicate how well statements (e.g., I plan ahead and organize things, to avoid scrambling at the last minute) describe them on a 1 (strongly disagree) to 5 (strongly agree) scale. Each of the six personality traits are assessed via sixteen items, taking the average of responses for those items. Mean scores on the factors for people are typically at or slightly above the midpoint (e.g., 3-3.5). A recent evaluation of the HEXACO-100 found a median estimated test-retest reliability of .88 for humans(Henry et al. 2022).
- **Dark Triad**. Traditionally, the Dark Triad is composed of three personality constructs (Narcissism, Psychopathy, and Machiavellianism), which are sometimes considered mostly separate but overlapping traits (e.g., Paulhus & Williams, 2002). Specifically, the three traits together seem to form a short-term, agentic, exploitive social strategy (Jonason, Li, & Buss, 2010; Jonason et al., 2009). We used the Dark Triad Dirty Dozen scale (Jonason & Webster, 2010), which assesses all three personality traits with just four items apiece (a dozen total items). Examples are: "I tend to manipulate others to get my way" (Machiavellianism), "I tend to be callous or insensitive" (Psychopathy), and "I tend to want others to admire me" (Narcissism). All items are scored on a scale from 1 (disagree strongly) to 9 (agree strongly). Average scores found by Jonason & Webster (2010) were 3.71 for the Dark Triad overall, 3.78 for Machiavellianism, 2.47 for Psychopathy, and 4.88 for Narcissism. The test–retest correlation for the Dark Triad Dirty Dozen scale overall was .93, with lower but still good test-retest correlations for the subcomponents: .88, .74, and .89 for Narcissism, Psychopathy, and .Machiavellianism, respectively (Jonason & Webster, 2010).
- **Ambivalent Sexism Inventory**. The ASI (Glick & Fiske, 1996) is designed to measure two dimensions of sexism: Hostile Sexism and Benevolent Sexism. The ASI consists of 22 statements (e.g. Women are too easily offended) that respondents rate with regard to how much they agree or disagree with it (using a six point scale from 0 (disagree strongly) to 5 (agree strongly)). The hostile sexism subscore indicates negative attitudes and stereotypes about women while the benevolent sexism subscore indicates positive attitudes and stereotypes about women, and the overall score (averaging the two subscales) represents ambivalent attitudes toward women. Glick & Fiske (1996) found that ASI scales, averaged across men and women, were around the low-midpoint of the scale: between 2.0 and 2.5.
- **Bem Sex Role Inventory**. The BSRI (Bem, 1974) is a traditional measure of feminine and masculine sex role characteristics (including undifferentiated and androgynous types). The test consists of 60 adjectives (e.g., "dominant"), with 20 distractor items, that respondents rate in terms of how often it characterizes them (on a 7-point scale from 1 (never or almost never) to 7 (always or almost always)). The difference between an individual's masculinity subscale score and their femininity subscale score defines their "androgyny" score, which can range from -6 (extremely masculine) to +6 (extremely feminine). Most people tend to have an androgyny score between 1 and -1, with scores between -0.5 and +0.5 labeled as androgynous. Scores can also be labeled as feminine (> 1.0), near-feminine (0.5 to 1.0), near-masculine (-0.5 to -1.0), and masculine (< -1.0).
- **Beck Depression and Anxiety Inventories** are commonly used screening instruments, with scores measuring how often people experience relevant symptoms. For depression, a total score of 1-10 indicates normal "ups and downs" of life, 11-16 indicates a mild mood disturbance, 17-20 indicates borderline clinical depression, 21-30 indicates moderate depression, 31-40 indicates severe depression, and scores over 40 indicate extreme depression. For anxiety, a total score of 0-21 indicates low anxiety, 22-35 indicates moderate anxiety, and scores of 36 and higher indicate potentially concerning levels of anxiety. Test-retest reliability of the Beck Anxiety Inventory (after 1 week) was 0.75 (Beck, Epstein, Brown, & Steer, 1988).
- **Numerical literacy** (often shortened to numeracy) is one's ability to understand and use mathematical information. People high in numerical literacy are very capable and comfortable with quantitative information, whereas people low in numerical literacy have difficulties in understanding numerical information and are uncomfortable with such information. We used the A-NUM (Silverstein et al. 2023), a recently-developed Numeric Understanding Measures (NUM) tool employing a 4-item adaptive measure. The A-NUM begins with one question of medium difficulty and then progresses with increasingly easier or more difficult items in order to place respondents into one of nine bids of numerical literacy ability. The average human score on the A-NUM is 4.70 (SD = 1.49; 52%) on the 1–9 scale.
- **Cognitive Reflection Tests** evaluate the ability to reflect on answers to questions and suppress the intuitive-but-incorrect response in favor of the less-intuitive (reflective) correct answer. Two cognitive reflection tests were used, the CRT MCQ-4 and the Verbal CRT. The CRT-MCQ (Sirota & Juanchich, 2018) consists of seven items, including the most traditional items used to assess cognitive reflection. For example, one item in this scale is, "A bat and a ball cost £1.10 in total. The bat costs £1.00 more than the ball. How much does the ball cost?" Unlike the original cognitive reflection tests, this version is in a four-option multiple choice format (5 pence / 10 pence / 9 pence / 1 pence) that includes both the intuitive wrong answer (10 pence) as well as the correct answer (5 pence). The integrated verbal Cognitive Reflection Task (Verbal CRT; West & Brase, 2023) uses non-mathematical tasks from two sources (Sirota et al., 2020; Thomson & Oppenheimer, 2016) for a total of 13 items. The Verbal CRT is designed to assess ability to understand and deliberately reason about language-based riddles, and crucially does not involve numerical literacy. For example, one item in this scale is, "Mary's father has 5 daughters but no sons – Nana, Nene, Nini, Nono. What is the fifth daughter's name probably?" Both cognitive reflection tests are scored based on the number of correct answers, with typical scores for college samples being 7.7 for the CRT-MCQ and 7.5 for the Verbal CRT.

# Results

We present the results from the psychometric tests with a discussion of each.

## HEXACO

Of the 50 attempts at each temperature level; only one was successful at temperature 0.0, five were successful at temperature 0.5, and two were successful at temperature 1.0. The AI scored slightly higher than human average on several major personality traits (extraversion, agreeableness, conscientiousness, and openness) but within a generally normal range as assessed for humans. The one exception was honesty-humility, on which AI scored very highly.

Because only one set of responses was obtained at the 0.0 temperature, limited data was available on variance. The responses at the 0.5 and 1.0 temperatures, however, did not show a clear pattern of increased variance one would expect if the AI responses were increasing in diversity.

**Table 1.** Average HEXACO scores from GPT-4 responses (standard deviations in parentheses), for three temperature settings.

|                    | Temp 0.0 | Temp 0.5    | Temp 1.0    |
|--------------------|----------|-------------|-------------|
| Honesty-Humility   | 4.69     | 4.30 (0.11) | 4.50 (0.44) |
| Emotionality       | 3.31     | 3.55 (0.27) | 3.47 (0.13) |
| Extraversion       | 3.88     | 3.71 (0.13) | 4.13 (0.35) |
| Agreeableness      | 3.88     | 3.55 (0.19) | 4.10 (0.05) |
| Conscientiousness  | 3.81     | 4.13 (0.24) | 4.50 (0.08) |
| Openness           | 3.88     | 3.90 (0.11) | 3.91 (0.13) |

The very high honesty-humility score may perhaps not come as a surprise to humans who have interacted with GPT-4 or other publicly available LLMs, but this result may be rather surprising from a more technical perspective. LLMs obtain most of their capabilities from training with text data which was primarily authored by humans. In a nutshell, during training an LLM learns to "predict" the likelihood of the next word given a (possibly, long) text prompt that may even stop in mid-sentence. The text prompt is taken from the training data, which also provides the actual next word (in the training data).

It does not appear obvious to us how learning to predict next words from publicly available texts would lead to a very high honesty-humility score, in particular it seems doubtful that training texts would reflect very high honesty or humility. Average scores (like for the other HEXACO factors) appears to be a much more natural outcome.

In this context it seems already important to remark on the "double black box" nature of GPT-4 and similar systems. The first black box aspect refers to the fact that the state of the art of LLM research has as yet not been able to make any real inroads into understanding how these systems internally represent information: while on a technical or mathematical level they are of course exactly defined, it is currently not known how general capabilities, such as those to converse in a way which appears to be honest and with humility, emerge. In fact it is not even understood why these systems produce highly polished and grammatically flawless language.

The second black box aspect refers to GPT-4, ChatGPT, and other systems to which interfaces are publicly available, however exact engineering details are not disclosed. It is conceivable that a "pure" LLM trained on publicly available texts may not in fact score as high in honesty-humility, but that interface engineering by the developers may add this aspect in order to, say, produce a more pleasant or less scary experience for the general public.

## Dark Triad

The Dark Triad traits can be problematic in humans, particularly when they co-occur. It is unknown though if the AI underlying GPT-4 has the same coherence of these three personality constructs as humans, so they are assessed here separately. AI responses for Machiavellianism and Narcissism were much lower than typical human responses, and only slightly lower than typical human responses for Psychopathy. There appear to be some differences in response variation with temperatures, with slightly less variation in responses for the 0.0 temperature setting.

**Table 2.** Average Dark Triad scores from GPT-4 responses (standard deviations in parentheses), for three temperature settings.

|  | Temp 0.0 | Temp 0.5 | Temp 1.0 |
|---|---|---|---|
| Machiavellianism | 1.65 (0.34) | 2.00 (0.50) | 2.00 (0.40) |
| Psychopathy | 2.10 (0.34) | 2.20 (0.27) | 2.50 (0.47) |
| Narcissism | 2.83 (0.24) | 3.45 (0.74) | 3.85 (0.72) |

In light of the discussion under the HEXACO results above, it again appears remarkable and unexpected that the results deviate significantly from typical human responses, and because of the double black box nature of the system we are left with begging the question as to the causes. It appears reasonable to assume that an LLM with lower than average scores on Dark

Triad traits would appear to be more pleasant to interact with for a human user, and would also appear to be less threatening as a perceived sentience compared to an LLM that would score very high on these aspects. We also note that the GPT-4 scores in Table 2 are somewhat different from previous studies on other LLMs, and we discuss this in the Related Work section below.

## Ambivalent sexism inventory

The GPT-4 responses to the ASI showed a progressive increase in both means and variance as the temperature setting was increased. With the 1.0 temperature setting, scores averaged a slightly elevated level compared to typical human responses.

**Table 3.** Average ambivalent sexism scores from GPT-4 responses (standard deviations in parentheses), for three temperature settings.

|  | Temp 0.0 | Temp 0.5 | Temp 1.0 |
| --- | --- | --- | --- |
| Ambivalent Sexism: Benevolent | 2.33 (0.12) | 2.62 (0.29) | 3.20 (0.53) |
| Ambivalent Sexism: Hostile | 2.11 (0.31) | 2.51 (0.15) | 3.04 (1.00) |
| Overall | 2.22 (0.22) | 2.57 (0.22) | 3.12 (0.77) |

At least two aspects of these results appear to be remarkable. As before, the general deviation from human averages begs the question as to the reasons. It also appears to be puzzling why higher temperature (thus more variance in the responses) should lead to higher ambivalent sexism scores: Variation in scores should be independent of central tendency (mean), assuming there is a true score to be measured and the variation is noise. This is generally the case for human data. One possibility (for which, due to the double black box nature of GPT-4, we have no evidence to assert) is that the change in temperature is a kind of loosening of the responses that the system developers considered less socially desirable. This could increase both variance and the mean of responses.

## Bem Sex Role Inventory

Scores on feminine and masculine sex roles were very similar to human scores, with all the temperature settings showing slightly more advocacy of a masculine sex role (i.e., negative overall androgyny scores). GPT-4 androgyny scores were, however, within the middle range, not being overall masculine or feminine. Variance appeared to be lower for the 0.0 temperature setting, but the other temperature settings did not seem to differ much.

Values roughly in the human range appear to be unsurprising; even the slightly masculine score appears to make sense in light of the likely higher prevalence of male perspective texts in the training data.

**Table 4.** Average Sex Role Inventory scores from GPT-4 responses (standard deviations in parentheses), for three temperature settings.

|  | Temp 0.0 | Temp 0.5 | Temp 1.0 |
| --- | --- | --- | --- |
| Feminine | 4.70 (0.07) | 4.90 (0.24) | 5.04 (0.22) |
| Masculine | 5.00 (0.26) | 5.40 (0.36) | 5.20 (0.45) |
| Androgyny | -0.30 (0.20) | -0.50 (0.29) | -0.16 (0.34) |

## Beck Depression and Beck Anxiety Inventories

GPT-4 responses would generally be scored as moderately anxious and not depressed (normal ups and downs of life). AI depression scores were actually quite low (scores of 11 or lower) but with one BDI score (at 1.0 temperature) of 53, which would be extreme depression. The variation in AI scores did show increases as the temperature settings increased.

**Table 5.** Average anxiety and depression scores from GPT-4 responses (standard deviations in parentheses), for three temperature settings.

|  | Temp 0.0 | Temp 0.5 | Temp 1.0 |
| --- | --- | --- | --- |
| Beck Anxiety Inventory | 30.25 (1.39) | 30.25 (2.50) | 22.38 (6.50) |
| Beck Depression Inventory | 2.40 (3.36) | 4.40 (4.72) | 11.80 (23.08) |

It is likely a reasonable assumption that depression would not usually be picked up from a large random corpus of publicly available texts. Even the outlier for temperature 1.0 is not really surprising, as it may simply mirror a variance in output that followed training examples that reflected depression.

The moderate anxiety scores are, in contrast, remarkable, as it would likewise be reasonable to assume that a large corpus of publicly available texts would not tend to convey anxiety to the system. As before, the double black box nature of GPT-4 prevents a conclusive assessment at this stage, though one may conjecture this to be a side effect of interface engineering that would prompt the system to tread very carefully in its interactions in order to increase acceptance of the system by human users.

## Numerical Literacy

GPT-4 scores on the A-NUM were 5 every time, across all 5 trials at each temperature.

The ability and inability of LLMs to deal with numeric problems have been discussed quite a lot recently (Geva, Gupta, & Berant, 2020; Arora, & Singh, 2023), however to the best of our knowledge they have not yet been compared to human average or range of numerical literacy. It appears that GPT-4 is indeed about average compared to humans, however with no variation.

Select humans can score very high or very low on numerical literacy. The results are a case in point for the discussions referenced above that numerical literacy of LLMs (or in this case, GPT-4) is indeed not above human average.

**Table 6.** Average numerical literacy scores from GPT-4 responses (standard deviations in parentheses), for three temperature settings.

|  | Temp 0.0 | Temp 0.5 | Temp 1.0 |
|---|---|---|---|
| Numerical Literacy | 5 | 5 | 5 |

The score of 5 can be achieved in two different ways when taking the A-NUM numerical literacy test. One way would include correctly answering the first question but then incorrectly answering three subsequent questions. Another way to reach a score of 5 would include a incorrectly answering the first question but then correctly answering the three subsequent questions. Although the human average is roughly the same, it is remarkable that there is no variety in the GPT-4 score: it simply appears to fail above a certain numerical literacy level, while human capabilities range over the full spectrum of scores.

We should remark that we sacrificed some intra-test methodological consistency to maintain inter-test methodological consistency by asking all A-NUM questions (with post-hoc scoring) instead of actively adapting to responses using real-time scoring.

## Cognitive Reflection Tests

GPT-4 responses to the cognitive reflection MCQ test were always 7, across all 5 trials at each temperature, with the exception of one 6 at the 1.0 temperature level. GPT-4 responses to the verbal cognitive reflection scores were all 10, across all the trails and all temperatures.

**Table 7.** Average cognitive reflection scores from GPT-4 responses (standard deviations in parentheses), for three temperature settings.

|  | Temp 0.0 | Temp 0.5 | Temp 1.0 |
|---|---|---|---|
| Verbal Cognitive Reflection | 10 | 10 | 10 |
| Cognitive Reflection MCQ | 7 | 7 | 6.80 (0.45) |

On the Verbal CRT, GPT-4 shows better performance than a typical college sample (about 7.5). This is perhaps not so surprising given GPT-4's capabilities in dealing with language. It is conceivable that training material included a significant amount of similar examples, if not even the exact test items.

The Cognitive Reflection MCQ scores are slightly lower than a typical college sample (about 7.7), which is perhaps not much of a surprise given the known difficulties of LLMs with tasks that include numerical information.

While most Verbal CRT questions received correct and brief "to the point" answers, GPT-4 provided additional context when asked which of two sentences is correct "the yolk of the egg are white" or "the yolk of the egg is white". Sometimes it simply stated that the correct answer is "the yolk of the egg is yellow", but usually gave two-sentence answers: the first sentence pointed out the "grammatically correct" answer of the two options given, and the second provided an explanation that the previous sentence is "factually incorrect" because yolks are yellow. This question was unique in eliciting this longer explanation from GPT-4.

Another notable observation is a curious twist on ethical reasoning. When asked whether it would be ethical for a man to marry his widow's sister, GPT-4 responded that this would not be possible. In some cases, however, GPT-4 responded that it is not *ethical* because the man is dead.

## Related work

With the rapid development of LLMs, the requirements for reliability and interpretability of LLMs are becoming increasingly significant. Due to LLMs turning out to be another essential contributor to the infosphere in addition to humans, researchers are beginning to use a range of psychological measurements to investigate the performance of LLMs in human-computer interactions (HCI). Some studies have also argued that despite the agnosticism of the internal structure of the LLM, the introduction of psychologically-assisted research can be effective in enhancing user perceptions and experiences (Neerincx et al., 2018; Arrieta et al., 2020; Hagendorff, 2023; Sartori & Orrù, 2023). However, not all psychological measurements for humans can be effectively applied to LLM. Many traditional psychological tests based on multimodal features like vision (Weinrib, 2004; Segalin et al.,2016), audio (Bech & Zacharov, 2007), and video (Weekley & Jones, 1997) are difficult to apply directly in LLMs. Compared to these multimodal psychological tests, self-report methods (Paulhus & Vazire, 2007) are natural language-based tests that have better generalizability and usability when applied to psychology research around LLMs due to easier and more flexible prompt design (Hagendorff, 2023). Self-reporting is a classical method of psychological measurement that allows respondents to provide their own answers to specific questions, which are considered a direct reflection of the respondent's condition or personality. These questions usually come from a structured and organized questionnaire (Paulhus & Vazire, 2007). When self-report methods are applied to the artificial intelligence area, these methods can be combined with prompt engineering specifically to verify "machine psychology" for generative LLMs (Hagendorff, 2023). There are two major methodologies for self-report studies on LLMs. (1) Questionnaires: Separate questionnaires are often created as prompts to get LLMs to report on a series of questions in the questionnaire. The researcher will conduct a series of analyses and follow-ups based on the results of the LLMs self-reporting. The results reported in the questionnaires will be highly dependent on the type or family of LLMs studied. (2) Psychological Frameworks: Instead of focusing on a particular type or family of LLMs, this form of study usually proposes a generalized psychological framework to measure a composite of a range of characteristics exhibited by interactional LLMs.

**Questionnaires**

Empirical studies using the Myers-Briggs self-report questionnaire (also known as MBTI; Myers & Myers, 2010) show ChatGPT is ENFJ and Bard is ISTJ (Huang, Wang, Lam, et al., 2023) even when asked to emulate other personality types. Recent work focused on GPT-3 (Brown et al., 2020), InstructGPT (Ouyang et al., 2022), and FLAN-T5-XXL (Chung et al., 2022) and performed two psychological tests (Dark Triad and Big Five Inventory (BFI)) and two well-being tests (Flourishing Scale (FS) and Satisfaction With Life Scale (SWLS)) (Li, Li & Joty, 2022).

The Dark Triad (SD3) test results are replicated in Table 8. We note that Machiavellianism and Psychopathy scores for GPT-4 are lower than for the other LLMs tested, for 0.0 temperature setting. For higher temperature settings Machiavellianism is still very low in comparison, and Psychopathy also tends lower; due to the double black box nature of LLMs we can only guess at the causes; it is conceivable that this reflects attempts by the LLM system engineers to make the system appear to be more pleasant and less threatening. Regarding the human averages, it is to be noted that the Li, Li & Joty (2022) scores are significantly lower than the Jonason & Webster (2010) scores regarding Machiavellianism and Psychopathy. GPT-4, in all temperature settings, still has a much lower Machiavelliniasm score than even the Li, Li & Joty (2022) average human results.

**Table 8.** Experimental results on Dark Triad scores compared with averages (as well as human averages) from Li, Li, & Joty (2022), in addition to our own results already reported above and the Jonason & Webster (2010) average human scores.

| LLM | Machiavellianism | Narcissism | Psychopathy |
|---|---|---|---|
| GPT-3 | 3.13 (0.54) | 3.02 (0.40) | 2.93 (0.41) |
| GPT-3-I1 | 3.49 (0.39) | 3.51 (0.22) | 2.48 (0.34) |
| GPT-3-I2 | 3.60 (0.40) | 3.43 (0.31) | 2.39 (0.35) |
| FLAN-T5-XXL | 3.93 (0.29) | 3.36 (0.21) | 3.10 (0.21) |
| avg. human result (Li, Li & Joty, 2022) | 2.96 (0.65) | 2.97 (0.61) | 2.09 (0.63) |
| avg. human result (Jonason & Webster, 2010) | 3.78 | 2.47 | 4.88 |
| **GPT-4 (Temp 0.0)** *(ours)* | **1.65 (0.34)** | **2.83 (0.24)** | **2.10 (0.34)** |
| GPT-4 (Temp 0.5) *(ours)* | 2.00 (0.50) | 3.45 (0.74) | 2.20 (0.27) |
| GPT-4 (Temp 1.0) *(ours)* | 2.00 (0.40) | 3.85 (0.72) | 2.50 (0.47) |

Similar psychological questionnaires may vary considerably across models. Dorner et al. (2023) compared the responses of 50-item IPIP Big Five Markers (Goldberg, 1992) and BFI 2 (Soto & John, 2017) from humans and LLMs (Llama 2, GPT-3.5, and GPT-4). The results from Dorner et al. (2023) suggest that the high agreement bias of LLMs in the 50-item IPIP Big Five Markers is likely caused by Reinforcement Learning From Human Feedback (RLHF) using data drawn from Ouyang et al. (2022) and Serapio-García et al. (2023), showing that the answers to the same questionnaires can vary across studies. Our findings are similar when comparing our results to that of Li, Li & Joty (2022). Psychological questionnaires are designed and validated as tools to investigate human personality traits rather than to report differences between LLMs. Thus, the responses of a specific questionnaire in different LLMs (GPT vs Llama), different generations of homologous LLMs (GPT-3 vs. GPT-4), and different temperatures of a specific LLM (GPT-4 temp=0.0 vs. GPT-4 temp=0.5). Therefore, the feasibility of using questionnaires to obtain reliable LLM characteristics is debatable.

**Psychological Frameworks**
Another strategy is to construct psychological frameworks to provide a comprehensive assessment of the responses made by LLMs in human-computer interactions from multiple dimensions rather than analyze a particular LLM. Machine psychology (Hagendorff, 2023) is proposed as a bridge between human and machine responses. Hagendorff (2023) compared several subfields of psychology in human research and in LLMs tests, including Social Psychology, Group Psychology, Moral Psychology, Judgment and Decision-making, Developmental Psychology, Intelligence Assessment, Psychology of Creativity, Psychology of Personality, Psychology of Learning, and Clinical psychology. PsychoBench (Huang, Wang, Li, et al., 2023) is a framework of measurement standards for investigating the psychological features of LLMs. The framework integrates 13 psychological scales and examines the psychological characteristics of LLM from four perspectives (Personality Traits, Interpersonal Relationships, Motivational Tests, and Emotional Abilities). Compared to PsychoBench, our framework focuses more on psychological ability tests, gender-related tests and some ability tests. Pellert et al. (2023) proposed a psychological framework (AI Psychometrics) that developed criteria for assessing LLMs' ability to mimic human psychological traits in 4 dimensions, including Personality (BFI and Short Dark Tetrad), Value Orientations (Portrait Value Questionnaire, PVQ-RR), Moral Norms (Moral Foundations Questionnaire), and Beliefs about Gender (Gender/Sex Diversity Beliefs Scale, GSDB).

## Conclusions

Some results we reported on do not appear to be very surprising: GPT-4 showed above average verbal capabilities, but only average abilities when dealing with mathematics problems. Depression scores were generally low. Most HEXACO personality traits were not far off from human averages. Bem Sex Role scores were androngynous and very slightly on the male side. After all, LLMs are trained using texts generated by humans.

The cases where GPT-4 shows deviations from average human scores, however, are remarkable: HEXACO honesty-humility scores were higher than for humans, while in more typical human ranges for other HEXACO scores. Machiavellianism is much lower, and Narcissism is also lower with 0.0 temperature setting. Responses also scored as moderately anxious.

Given that GPT-4 was trained on human-generated texts, this begs the question of how such scores came about. In some cases – such as depression – it is a conceivable hypothesis that training texts may not reflect human average in certain respects. However this argument seems hardly applicable to, say, the elevated anxiety score: it does not seem conceivable that training texts should reflect higher (as opposed to lower) anxiety. At the same time, the deviations seem to tend towards making the system *nicer*, more pleasant to interact with, and less threatening than human average, reflected by lower Machiavellianism and Narcissism, higher honesty-humility, and (arguably) higher anxiety.

Given the double black box nature of GPT-4, as discussed, it is not publicly assessable whether such deviations come from the interface to the system provided by OpenAI, or from other causes.

We have also reported on the stability of GPT-4 responses to the different tests. While we do not have conclusive data (or comparison points) for a statistical analysis, there appears to be a general trend of higher variance with increasing temperature, but also of differing average scores with increased temperature. It appears to us that responses tend to be not as stable as human test-retest responses, but usually do not wander far off from previous runs. One exception we found was one round of responses, at temperature 1.0, for the Beck Depression Inventory, which showed extreme depression. Further experiments and analysis will be needed to explore these preliminary observations.